\documentclass[runningheads]{llncs}

\usepackage{eccv}

\usepackage{eccvabbrv}

\usepackage{graphicx}
\usepackage{booktabs}

\usepackage{multirow}
\usepackage{amssymb}
\usepackage{amsmath}
\usepackage{stfloats}
\usepackage{colortbl}
\usepackage{float}
\usepackage[misc]{ifsym}
\usepackage{soul, color, xcolor}

\makeatletter

\newcommand{\Rmnum}[1]{\expandafter\@slowromancap\romannumeral #1@}
\makeatother

\usepackage[accsupp]{axessibility}  

\usepackage{hyperref}

\usepackage{orcidlink}

\begin{document}

\title{MO-EMT-NAS: Multi-Objective Continuous Transfer of Architectural Knowledge Between Tasks from Different Datasets} 

\titlerunning{MO-EMT-NAS}

\author{Peng Liao\inst{1}\orcidlink{0009-0006-9711-7142} \and
XiLu Wang\inst{3}\orcidlink{0000-0002-0926-4454} \and
Yaochu Jin\inst{1,2} \textsuperscript{(\Letter)} \orcidlink{0000-0003-1100-0631}\and
WenLi Du\inst{1}\textsuperscript{(\Letter)} \orcidlink{0000-0002-2676-6341}}

\authorrunning{P.Liao, X.Wang et al.}

\institute{Key Laboratory of Smart Manufacturing in Energy Chemical Process, Ministry of Education, East China University of Science and Technology, China \\
\email{pengliao@mail.ecust.edu.cn, wldu@ecust.edu.cn} \\
Trustworthy and General AI Lab, School of Engineering, Westlake University, Hangzhou, 310030, PR China\\
\email{jinyaochu@westlake.edu.cn}\\
Computer Science, University of Surrey, Surrey, GU2 7XH, UK\\
\email{wangxilu@surrey.ac.uk}
}

\maketitle

\begin{abstract}
    Deploying models across diverse devices demands tradeoffs among multiple objectives due to different resource constraints. Arguably, due to the small model trap problem in multi-objective neural architecture search (MO-NAS) based on a supernet, existing approaches may fail to maintain large models. Moreover, multi-tasking neural architecture search (MT-NAS) excels in handling multiple tasks simultaneously, but most existing efforts focus on tasks from the same dataset, limiting their practicality in real-world scenarios where multiple tasks may come from distinct datasets. To tackle the above challenges, we propose a Multi-Objective Evolutionary Multi-Tasking framework for NAS (MO-EMT-NAS) to achieve architectural knowledge transfer across tasks from different datasets while finding Pareto optimal architectures for multi-objectives, model accuracy and computational efficiency. To alleviate the small model trap issue, we introduce an auxiliary objective that helps maintain multiple larger models of similar accuracy. Moreover, the computational efficiency is further enhanced by parallelizing the training and validation of the weight-sharing-based supernet. Experimental results on seven datasets with two, three, and four task combinations show that MO-EMT-NAS achieves a better minimum classification error while being able to offer flexible trade-offs between model performance and complexity, compared to the state-of-the-art single-objective MT-NAS algorithms. The runtime of MO-EMT-NAS is reduced by 59.7\% to 77.7\%, compared to the corresponding multi-objective single-task approaches.
  \keywords{Neural architecture search (NAS) \and Multi-objective optimization (MO) \and Multi-task learning \and  Parameter sharing }
\end{abstract}

\section{Introduction}
\label{sec:intro}
EMT-NAS \cite{Liao_2023_CVPR} is a recently proposed multi-tasking NAS algorithm that aims to address the challenge of multiple tasks from different datasets. Different from the initial shared representation of tasks from the same dataset, EMT-NAS considers knowledge transfer from one task to a related task, e.g., transferring knowledge from playing squash to playing tennis \cite{zhang2018overview}. EMT-NAS ensures that each task has a separate set of supernet parameters, skillfully alleviating the negative transfer \cite{vandenhende2021multi} that may result from joint training of weight parameters for multiple tasks. However, it evaluates the architectural accuracy only and therefore tends to favor larger models in the search space, lacking control over the model size. To address the above limitations, the present work aims to identify a set of architectures that can balance multiple objectives for each task by means of multi-objective optimization (MO), successfully tackling the complexities posed by multiple classification tasks from diverse datasets.

Many existing studies on single-task NAS have adopted the MO approach to strike a trade-off between the accuracy and other indicators required in a wide range of settings, such as computational complexity, CPU and GPU latency \cite{NEURIPS2021_e3251075}, adversarial robustness \cite{guo2020meets}, and data privacy \cite{9664283}. Evolutionary MO-NAS based on the weight-sharing-based supernet has achieved remarkable success, which significantly reduces computational consumption by allowing all possible architectures to share parameters. However, it has been found that when considering the deployment of models on diverse devices with varying resources, simultaneously optimizing the classification error and model size often drives the population to quickly converge to smaller models. This phenomenon arises because smaller models converge faster in the early stage of evolutionary optimization \cite{yu2019evaluating}, resulting in lower classification errors for small models \cite{ren2021comprehensive}. Hence, environmental selection based on the non-dominance relationship favors smaller models and eliminates larger ones. To address the above issue, CARS \cite{yang2020cars} designed two environmental selection strategies, one minimizing the model error and size that favors small models, and the other minimizing both the model error and convergence speed that favors larger models. It was found, however, that the population was able to maintain a good degree of diversity in the initial stages; however, the population became polarized into large and small models as the evolution proceeded, losing the models of medium sizes. 

To tackle the challenges of multi-objective MT-NAS, this work adopts multi-objective multi-factor evolutionary algorithms (MO-MFEAs) \cite{gupta2016multiobjective} to transfer latent similarity knowledge across different tasks. Furthermore, an auxiliary objective is introduced to maintain the architecture diversity so that small, medium, and large models can be retained, ensuring that the final solution set contains trade-off models of a wide range of model sizes. 
Key contributions are as follows:

\begin{itemize}
\item We propose an MO-EMT-NAS framework to effectively search for multiple Pareto optimal architectures for tasks from different datasets, utilizing transferable architecture knowledge across the tasks to facilitate continuous architecture search.
\item MO-EMT-NAS considers both the classification error and model size, and the auxiliary objective that mitigates the search bias towards small neural architectures. With the help of multiobjectivization, MO-EMT-NAS maintains architecture diversity in terms of model sizes. 
\item The computational efficiency of MO-EMT-NAS is improved by allowing parallel training and validation of the task-specific weight-sharing supernet on each task.
\item We benchmark MO-EMT-NAS's performance on seven datasets with two, three, and four tasks, respectively. These datasets include CIFAR-10, CIFAR-100, ImageNet, and four medical datasets (PathMNIST, OrganMNIST\_\{Axial, Coronal, Sagittal\}). From the results, we see that MO-EMT-NAS can obtain architectures with trade-offs between the performance and model size. Through implicit architectural knowledge transfer across different tasks, MO-EMT-NAS can achieve better-performing neural architectures while using less runtime compared to the multi-objective single-tasking approach.
\end{itemize}

\section{Related Work}
Neural architecture search aims at automatically finding neural network architectures that are competitive with those manually designed by human experts \cite{MoblieNet-V2,ding2021repvgg,ShuffleNet,howard2019searching,hou2021coordinate, zhou2020rethinking}. Reinforcement learning (RL) \cite{ENAS}, gradient descent (GD) \cite{xie2018snas,liu2018darts}, and evolutionary algorithms (EA) \cite{wang2022continuous,AE-CNN} are three typical search strategies used in NAS. Most NAS algorithms, however, suffer from huge computational costs \cite{liu2022NAS}, leading to the development of one-shot NAS that can significantly reduce GPU days and lower the high demand for computational resources through parameter sharing \cite{pham2018efficient}.

{\noindent\bf Multi-objective NAS} has been developed to search for neural network models for optimizing objectives in addition to the accuracy required in real-world applications. Among the existing MO-NAS approaches, evolutionary MO, a population-based method, has been widely adopted as it is capable of achieving a set of Pareto optimal neural architectures in a single run. NASGNet \cite{lu2019nsga} was proposed to generate a set of architectures by simultaneously minimizing classification error and model complexity (floating-point operations per second, FLOPs) with a Bayesian network as a performance predictor. Alternatively, MT-ENAS \cite{9504721} adopted the network performance and model size as two objectives and used multi-task training to construct a radial-basis-function neural network \cite{liao2020multi} as a performance predictor. It is worth noting that they utilized separate populations for each objective without knowledge transfer. In NSGANetV2 \cite{lu2020nsganetv2}, five objectives are simultaneously optimized with the help of multiple performance predictors. 
Since most MO-NAS approaches rely heavily on the quality of the performance predictors, we demonstrate the differences between MO-EMT-NAS and them in that MO-EMT-NAS performs training and validation based on a weight-sharing supernet to reduce the computational overhead, instead of using performance predictors \cite{guo2023pareto} or zero-shot metric evaluation \cite{lin2021zen}. Although the use of parameter sharing allows candidate submodels to be easily evaluated without training from scratch \cite{real2017large}, candidates with small model sizes generally achieve better validation accuracy at the beginning of the search. Hence, promising candidate models with large sizes fail to survive to the next generation, resulting in a search bias towards small models.

{\noindent\bf Multi-tasking NAS:} NAS has evolved from single-task and transfer learning to multi-task learning, with the latest search focusing on different datasets. Since the NAS algorithm \cite{zoph2017neural} was proposed, NAS has demonstrated much success in automatically designing effective neural architectures. Initially, it was employed for optimizing models for specific single tasks, such as the classification task on CIFAR-10 \cite{krizhevsky2009learning}, termed single-task learning. As related tasks can be encountered, researchers resorted to transfer learning to improve NAS by transferring knowledge from similar previous tasks. For example, a pre-trained model is employed to guide the search for a new task \cite{liu2021transtailor, liu2022data}. Meanwhile, it has been observed that different tasks can stem from the same dataset. For example, instance and semantic segmentation and depth prediction can be performed on a large dataset for road scene understanding, CityScapes \cite{cordts2016cityscapes}. By learning shared representations across these tasks, a common neural architecture for all tasks is constructed via multi-task learning, instead of searching for a task-specific model for each task in the traditional approach \cite{NEURIPS2020_634841a6, Gao_2020_CVPR}. Regardless of its well-known efficiency, this line of research is limited to considering multiple tasks from the same dataset \cite{caruana1997multitask}. Recently, the presence of tasks from different datasets poses challenges for multi-tasking NAS, due to the fact that two tasks from different datasets show lower relatedness scores compared to those originating from the same dataset \cite{khattar2021cross}. Arguably, EMT-NAS \cite{Liao_2023_CVPR} was first developed as an MT-NAS with the help of an evolutionary multi-tasking framework to address tasks on different datasets.

Although recently proposed methods have shed light on the advantages of incorporating transfer learning and multi-task learning into NAS, our work establishes a multi-objective multi-tasking framework and focuses on handling multiple tasks on different datasets and providing a set of Pareto optimal architectures by balancing the model error and model size.


\section{Approach}
\label{sec:approach}

In this work, we used the search space of~\cite{zoph2018learning}. The encoding of a neural architecture consists of normal and reduction cells, with each cell comprising five blocks, and each block containing two input bits and two operation bits, amounting to a total of 40 bits. For each operation bit, candidate operators include depthwise-separable convolution, dilated convolution, max pooling, average pooling and identity. More details can be found in Supplementary Material A.

\subsection{MO-EMT-NAS}

\begin{figure}[!t]
   \centering
   \includegraphics[width=1.00\linewidth]{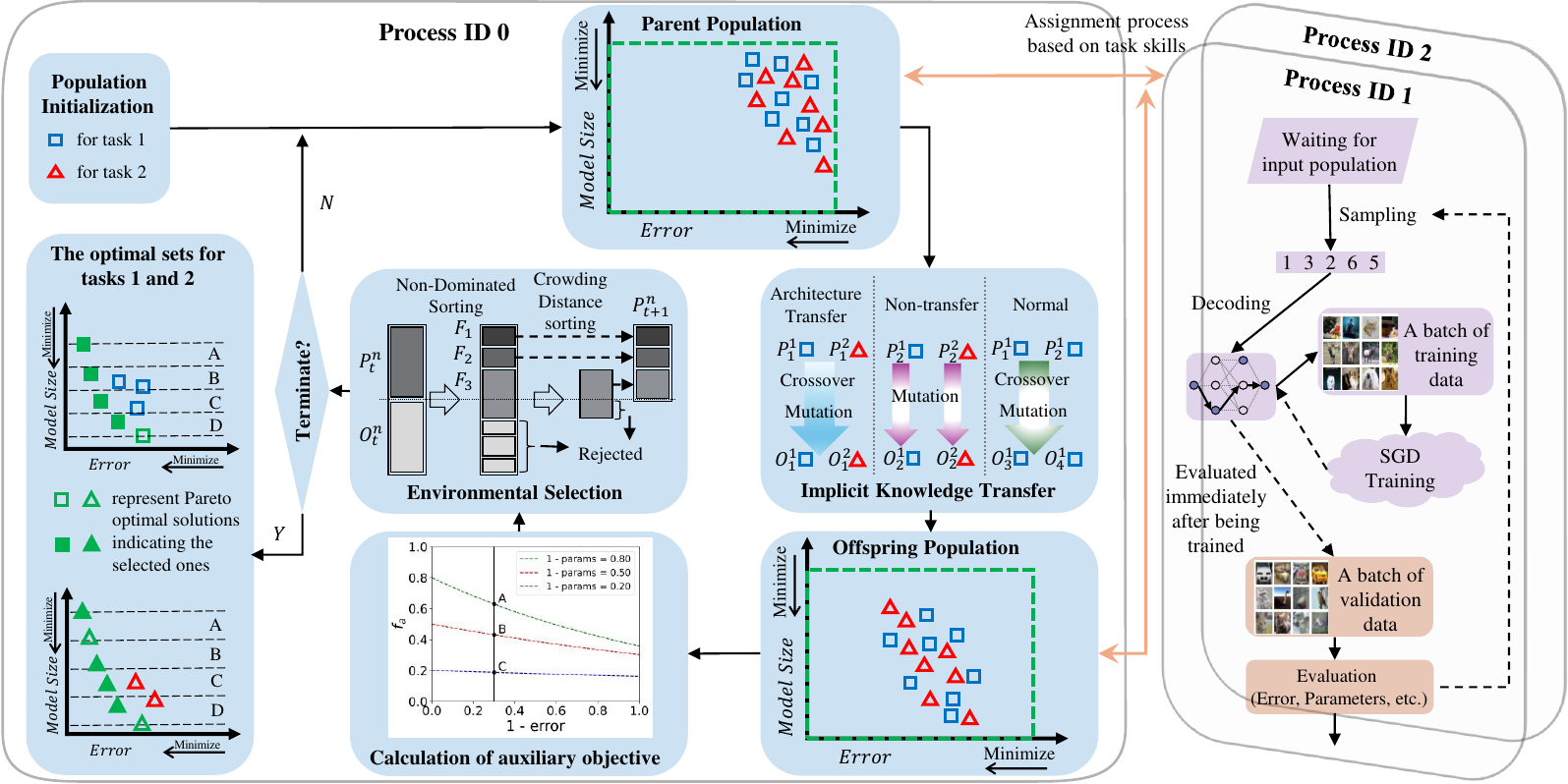}
   \caption{A flowchart of MO-EMT-NAS, using a two-task scenario as an example. Different from single-task multi-objective evolutionary approaches, in MO-EMT-NAS, individuals in a population belong to different tasks, and cross-task reproduction enables implicit knowledge transfer. To alleviate negative transfer, each task operates with its own supernet and the corresponding parameter set. Hence, implementing the concept of multiprocessing naturally separates the optimization algorithm from each task's training and validation, further enhancing the computational efficiency.}
   \label{fig:MO-EMT-NAS}
\end{figure}

The overall framework of MO-EMT-NAS is shown in Fig.~\ref{fig:MO-EMT-NAS}, including two main parts: the search algorithm and the training and validation of the weight-sharing-based supernet. Implementing multiprocessing enables parallelized training and validation of the supernet for various tasks, significantly boosting computational efficiency. 
See Supplementary Material B for more details and the pseudocode.

The architecture search algorithm is shown on the left panel of Fig.~\ref{fig:MO-EMT-NAS}. First, individuals of the initial population are randomly assigned to different tasks as the corresponding parent population. The parent individuals are sampled from the supernet and trained on each task, and then their objectives are evaluated. Then, the main loop is performed as follows. Individuals are selected (called mate selection) from the parent population to generate an offspring population by exploring the same tasks and transferring knowledge across different tasks. Both offspring and parent individuals are sampled from the supernet and trained on each task and then their objectives, i.e., the model error, size and the auxiliary objective, are evaluated. The non-dominated sorting with the crowding distance is performed on the combined parent and offspring population to select the population for the next generation. After repeating the main loop for several generations, a set of Pareto optimal solutions for each task is obtained.


In this work, block-based crossover and bit-based mutation of~\cite{Liao_2023_CVPR} are adopted due to the discrete coding of NAS. The block-based crossover uses the block as the basic unit and allows the selected two parents to exchange blocks at a predefined crossover probability. Bit-based mutation adopts bits as the basic unit of bits and randomly varies the bits of a selected single parent within a candidate range at the mutation probability. Within the framework of MO-EMT-NAS, the generation of offspring populations enables implicit knowledge transfer between tasks: 1) For parents assigned to the same task, the offspring is generated via crossover, and mutation operators can explore the corresponding task. When parents come from different tasks, the generation of offspring is controlled by a parameter called random mating probability (RMP). 2) Architectural knowledge transfer is triggered at a probability of $RMP$, where the offspring are generated from the parents through crossover and mutation, and are assigned to the tasks as one of its parents. 3) otherwise, no knowledge transfer will happen, i.e., the parents independently undergo mutation to generate the corresponding offspring, and these offspring inherit the task of their parents.  

In MO-EMT-NAS, environmental selection must consider multiple conflicting objectives on multiple tasks. Accordingly, the population should be divided into subpopulations by equipping individuals with different tasks. Thus each task can execute its environmental selection separately. Subsequently, a multi-objective environmental selection is performed to consider the validation error, the number of model parameters, and the auxiliary objective, aiming to enhance the architecture diversity and provide a set of promising architectures.

\subsection{Auxiliary Third Objective}
\label{sec:Auxiliary Third Objective}
 

Consider a minimization problem with $M$ objectives, individual $A$ dominates individual $B$, i.e., $A$ is better than $B$, if:
\begin{equation}
\begin{aligned}
f_m (A) \leq  f_m (B), \quad \forall m \in \{1, 2, \ldots, M\},    \\  
f_m (A) < f_m (B),  \quad \exists \, m \in \{1, 2, \ldots, M\}. 
\end{aligned}
\end{equation}
If $A$ does not dominate $B$ and $B$ does not dominate $A$, then $A$ and $B$ are non-dominated to each other, indicating $A$ and $B$ are similar. Similarly, $A$ is dominated by $B$ means that $A$ is worse than $B$. The selection of non-dominated solutions in NSGA-II~\cite{deb2002fast} is outlined as follows and depicted in Fig.~\ref{fig:MO-EMT-NAS}: 1) Non-dominated sorting is performed on the combined population of the parent and offspring, resulting in a non-dominated level/rank for each individual. 2) A predefined number of individuals are selected to survive to the next generation based on their non-dominated level. 3) If the number of individuals at the last accepted level exceeds the predefined population size, the crowding distance of each individual (indicating its contribution to solution diversity) is used as the selection criterion. Solutions with a large crowding distance will be prioritized to ensure the diversity of the population. 



\begin{figure}[!t]  
\centering
\subfloat[][Two objectives]{\includegraphics[width=1.2in]{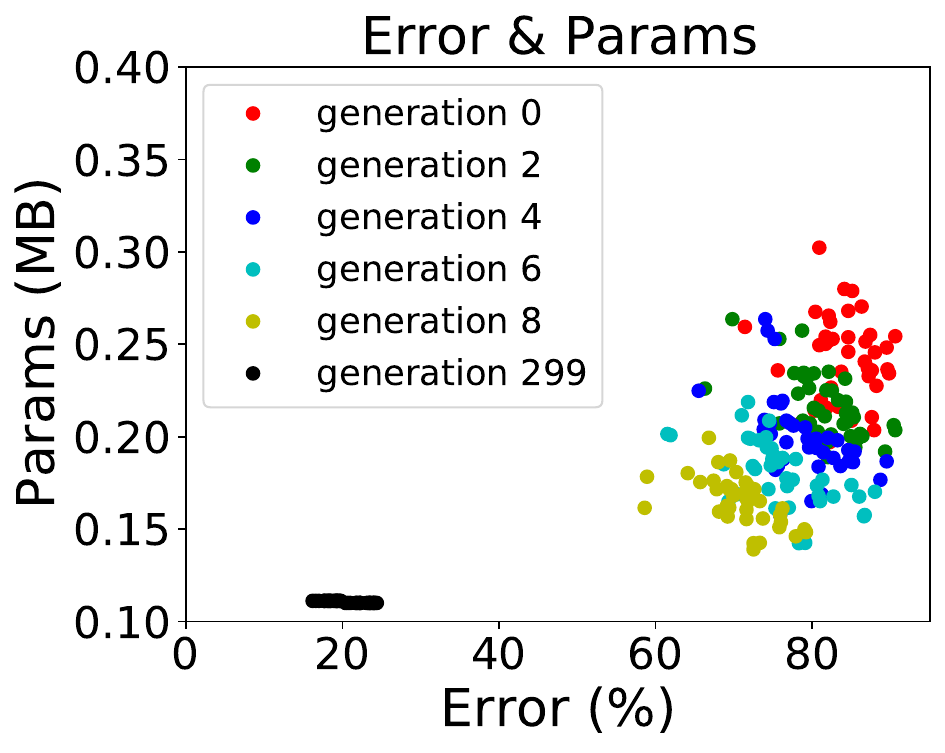}%
\label{function a}}
\subfloat[][MACs as the third objective]{\includegraphics[width=1.2in]{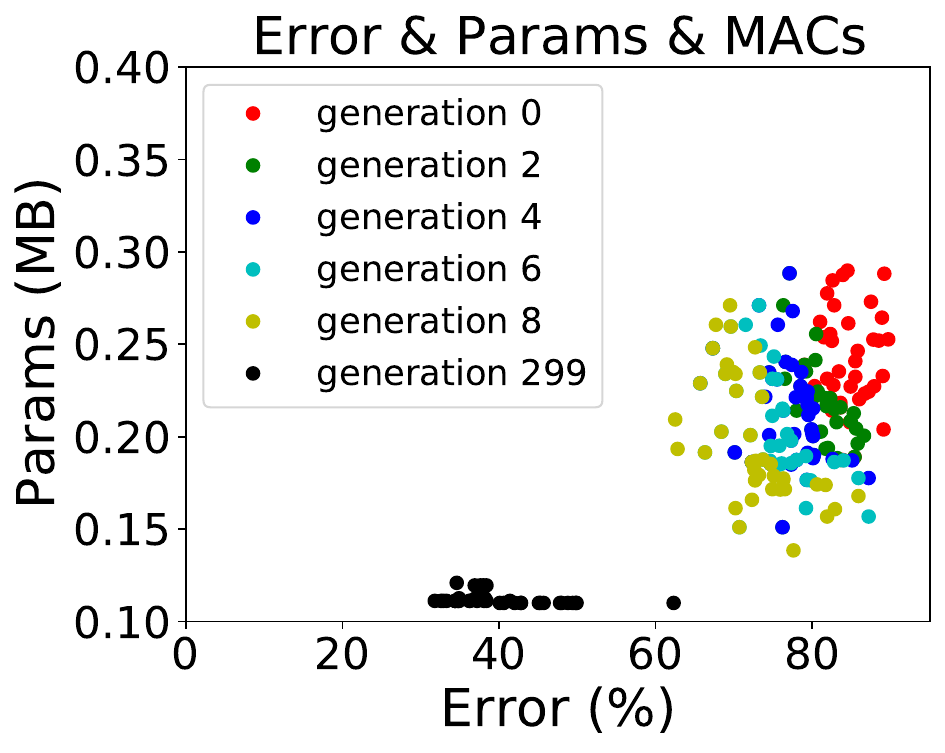}%
\label{function b}}
\subfloat[][The speed of increasing accuracy]{\includegraphics[width=1.2in]{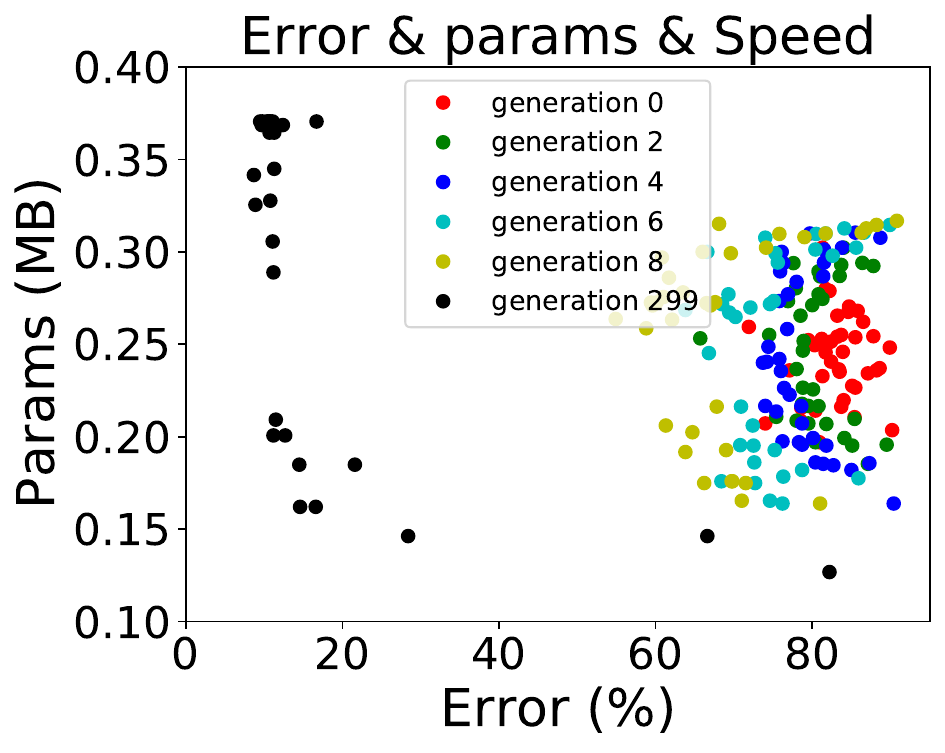}%
\label{function c}}
\subfloat[][$f_a$ as the third objective (Ours)]{\includegraphics[width=1.2in]{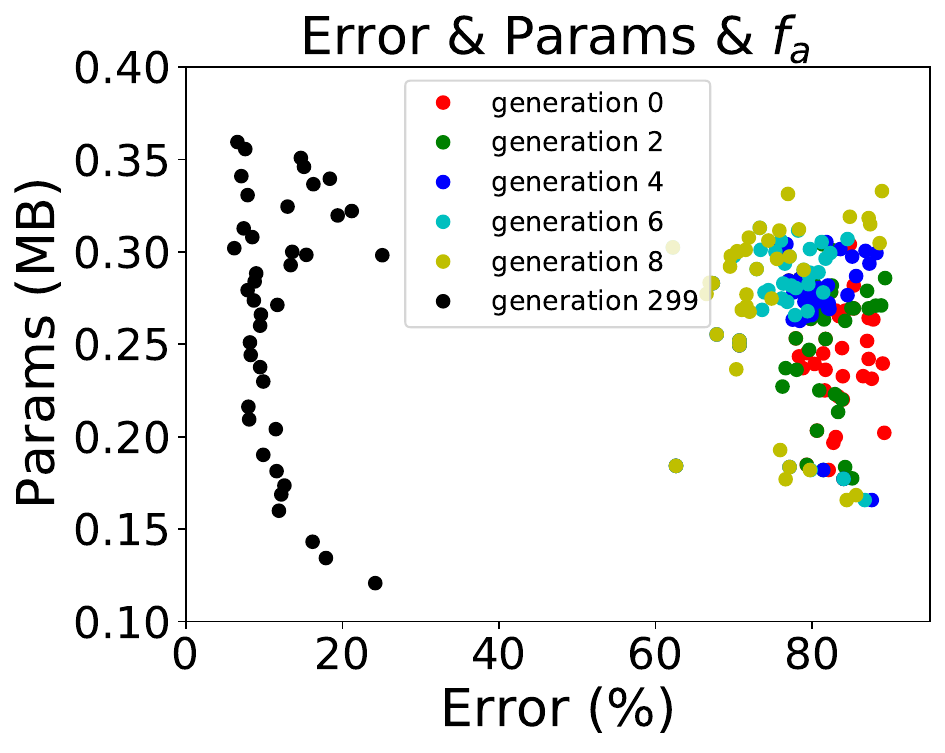}%
\label{function d}}
\caption{Comparison of the impact of different evolutionary strategies on the convergence of the population, each showing the population distribution in the initial, second, fourth, sixth, eighth, and the final generations. By simultaneously optimizing the model error, size and the proposed auxiliary objective, we can obtain models of various sizes.}
\label{3th ED}
\end{figure}

The key insight the multi-objective NAS approach can offer is to provide a bigger picture of the trade-offs between multiple important objectives for a real-world application. Unfortunately, achieving a set of diverse and promising architectures to simultaneously minimize validation error and model size is non-trivial, as previously discussed. Figure~\ref{3th ED} visualizes the populations obtained by NSGA-II with different objectives at different generations. In Fig.~\ref{function a}, the population obtained by minimizing both the model error and size converges rapidly towards small-size models as the evolution proceeds. This can be attributed to the fact that small-size models can achieve a superior validation error in the initial phase when using a weight-sharing-based supernet, resulting in losing all large models in the final population. To mitigate this issue, a practical approach is to include an auxiliary extra objective for improving the diversity of candidate architectures when performing non-dominated sorting. Figure~\ref{function b} depicts the results of adding the multiply-accumulate operations (MACs) as a third auxiliary objective. Unfortunately, small models still dominate large ones, simply because MACs provide a similar selection pressure to the model size. Alternatively, CARS \cite{yang2020cars}, a state-of-the-art NAS method, introduced a third objective, called accuracy speed, which is measured by the reciprocal of the number of model parameters. CARS performs the non-dominated sorting twice at each generation, one considering the validation accuracy and the number of model parameters, and the other the accuracy and the accuracy speed. As a result, CARS can maintain small and large models, but cannot retain medium-size ones, as shown in Fig.~\ref{function c}.

To resolve the above issue, this work introduces an auxiliary objective $f_a$ to maintain large models in the population by integrating both the model accuracy and size, and utilizing exponential distributions with respect to the size. Specifically, $f_a$ is defined as follows: 
\begin{equation}
f_a =(1-params)e^{-(1-params)(1-error) },
\label{eq:f_a}
\end{equation}
where the number of model parameters $params$ and the validation error $error$ are normalized to $[0,1]$ across the population of the current generation. The maintenance of large models is achieved by generating different exponential distributions \cite{wang2023alleviating} with respect to $1-params$. According to Eq.(\ref{eq:f_a}), a larger model will result in a smaller value of $1-params$ and accordingly a smaller $f_a$, compared to that of a smaller model with a similar $error$. Figure \ref{fig:MO-EMT-NAS} includes an example of the calculation of the auxiliary objective $f_a$: three architectures $A$, $B$, and $C$, achieve the same error of $0.7$, i.e., $1-error=0.3$. As a result, $f_a$ of $C$ with $1-params=0.2$ will be smaller than that of $A$ with $1-params=0.8$. Meanwhile, $f_a$ guides the search towards minimizing the validation error, since $f_a$ always prefers models with a lower $error$. Therefore, $f_a$ will not only strike a balance between $error$ and $params$, but also alleviate the search bias by carefully prioritizing the selection of large models. Therefore, the diversity of architectures can be enhanced by employing $f_a$ as the auxiliary objective, resulting in a more even distribution of model sizes in the population, as shown in Fig.~\ref{function d}.

\subsection{Parallel Training and Evaluation}

The training and validation of the weight-sharing-based supernet is shown on the right panel of Fig.~\ref{fig:MO-EMT-NAS}. The multi-tasking framework allows the training and validation of the supernet for each task to be parallelized. Specifically, the available number of iterations is divided by the number of individuals to obtain the number of iterations $train_n$ for each individual. This process utilizes only one epoch of training data. Following this, individuals of each task are decoded as a neural architecture and trained for $train_n$ times. Finally, the validation error and the number of model parameters of the population of each task are obtained. This way, each neural architecture can substantially enhance its validation accuracy during the successive training iterations, effectively reflecting its true performance \cite{yu2021analysis}.

\section{Experiments}
\label{sec:experiments}

\subsection{Settings}

We adopt a Multi-Objective Single-Tasking NAS (MO-ST-NAS) baseline by only removing the multi-tasking setting from MO-EMT-NAS to show the promising advantages of transferring architectural knowledge across related tasks on diverse datasets. Similarly, MO-EMT-NAS is compared with a representative single-objective evolutionary MT-NAS (EMT-NAS~\cite{Liao_2023_CVPR}) to demonstrate the efficiency of MO with the auxiliary objective. 

We select seven datasets and conduct experiments on two, three and four tasks for performance evaluation. 1) We design a two-task experiment on the classical datasets CIFAR-10 and CIFAR-100 \cite{krizhevsky2009learning}. Additionally, the obtained architectures are retrained on ImageNet \cite{ILSVRC-2012} in order to examine our MO-EMT-NAS's transferability. 2) Multiple tasks, i.e., two-, three- and four-task settings, are designed on MedMNIST \cite{yang2021medmnist} to validate the generalization ability of our method. Four datasets, namely PathMNIST, OrganMNIST\_Axial, OrganMNIST\_Coronal, and OrganMNIST\_Sagittal, are selected to simulate various medical imaging scenarios such as rectal cancer pathology and 2D images from 3D computed tomography (CT) images of liver tumors in different planes. Both our baseline and proposed approach are performed five independent runs, and the hyperparameters are listed in Table~\ref{settin}. 
See Supplementary Material C for more experimental setups.

Following the practice in~\cite{yang2020cars,wang2022continuous}, to better visualize and compare the optimal architectures obtained from the MO algorithms, the final population is divided into four groups based on the model size as shown in the example given in Fig.~\ref{fig:MO-EMT-NAS}, and the architecture with the smallest error is selected from each group (denoted as A, B, C, and D).

\begin{table}[!t]
    \centering
    \begin{minipage}[c]{0.79\linewidth}
\scriptsize
\caption{Summary of Hyperparameter Settings.}
\label{settin}
\centering
\begin{tabular}{ccc}
\hline
\multirow{1}{*}{Categories}   & Parameters          & Settings (CIFAR/MedMNIST)      \\ 
\hline
\multirow{3}{*}{Search space}     & Initial channels                     & 20/48            \\
                                  & Cell repetitions in the search & 1              \\
                                  & Cell repetitions in testing & 3/1              \\
\hline
\multirow{8}{*}{Gradient descent} & Batch size for training        & 128        \\
                                  & Batch size for validation   & 1000           \\
                                  & Weight decay                 & 1.00E-04         \\
                                  & Dropout                   & 0.1 \\
                                  & Generation start of dropout & 50 \\
                                  & Momentum                     & 0.9              \\
                                  & Initial learning rate        & 0.1              \\
                                  & Learning rate schedule       & Cosine Annealing \cite{loshchilov2016sgdr}  \\
\hline
\multirow{6}{*}{Search strategy}  & Population size              & 80               \\
                                  & Generations (Epochs)                 & 300/100              \\
                                  & Random mating probability    & 1.00              \\
                                  & Crossover probability        & 1.00             \\
                                  & Mutation probability         & 0.025             \\
                                  & Solution set size per task            & 4           \\
\hline
\end{tabular}
    \end{minipage}
    \hfill
    \begin{minipage}[c]{0.19\linewidth}
        \centering
        \includegraphics[width=0.95\linewidth]{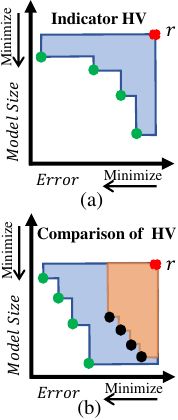}
        \captionsetup{type=figure}
        \caption{Illustration of HV.}
        \label{fig:HV_illustration}
        \vspace{-0.4cm}
    \end{minipage}
\end{table}

\subsection{Performance Indicator}
We adopt hypervolume (HV) \cite{zitzler1999multiobjective} to evaluate the sets of architectures found by different approaches in terms of convergence and diversity. HV calculates the volume of the objective space dominated by a set of non-dominated solutions $\mathcal{P}$ and bounded by a reference point $\mathbf{r}$ (see Fig.~\ref{fig:HV_illustration}a),
\begin{equation}
\operatorname{HV}(\mathcal{P})=VOL\left(\cup_{\mathbf{y} \in \mathcal{P}}[\mathbf{y}, ]\right),
\end{equation}
where $VOL(\cdot)$ denotes the usual Lebesgue measure, $[\mathbf{y}, \mathbf{r}]$ represents the hyper-rectangle bounded by $\mathbf{y}$ and $\mathbf{r}$. A larger HV value means better performance: In Fig. \ref{fig:HV_illustration}b, the set of converged and well-distributed green dots, exhibiting a higher HV value, achieves better performance compared with the set of black dots. For each task, after each separate run of all compared algorithms, the maximum values of each objective across all solutions form the reference point $\mathbf{r}$. Therefore, $\mathbf{r}$ varies across different tables that involve different algorithms.

\subsection{Two-task on CIFAR-10 and CIFAR-100}

\begin{figure}[!t]
\centering
\subfloat[][Task 1 (CIFAR-10)]{\includegraphics[width=0.50\textwidth]{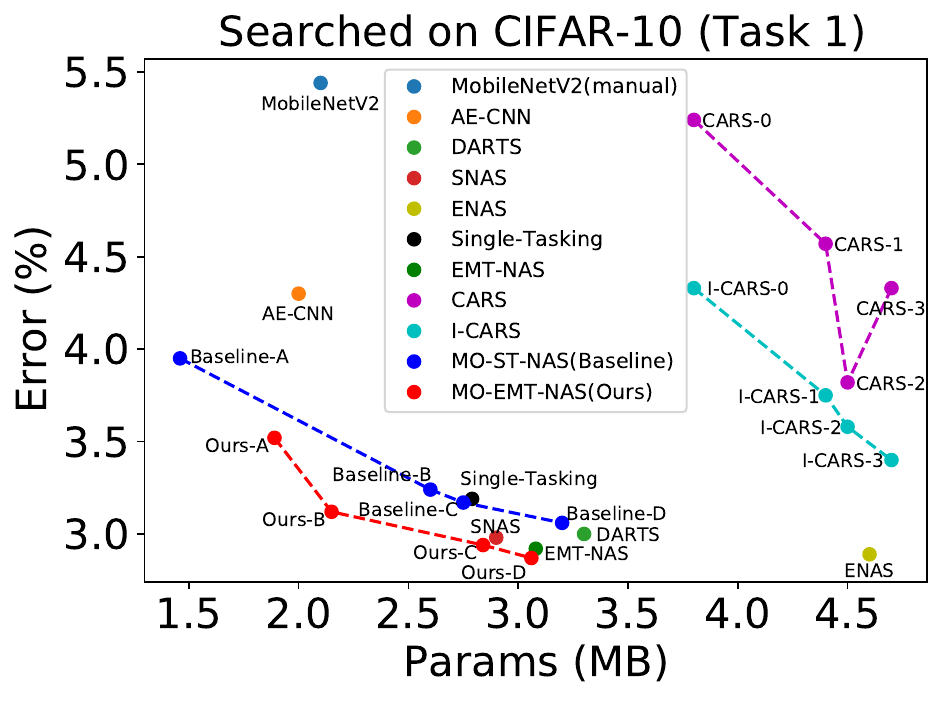}%
\label{CIFAR-10}}
\subfloat[][Task 2 (CIFAR-100)]{\includegraphics[width=0.50\textwidth]{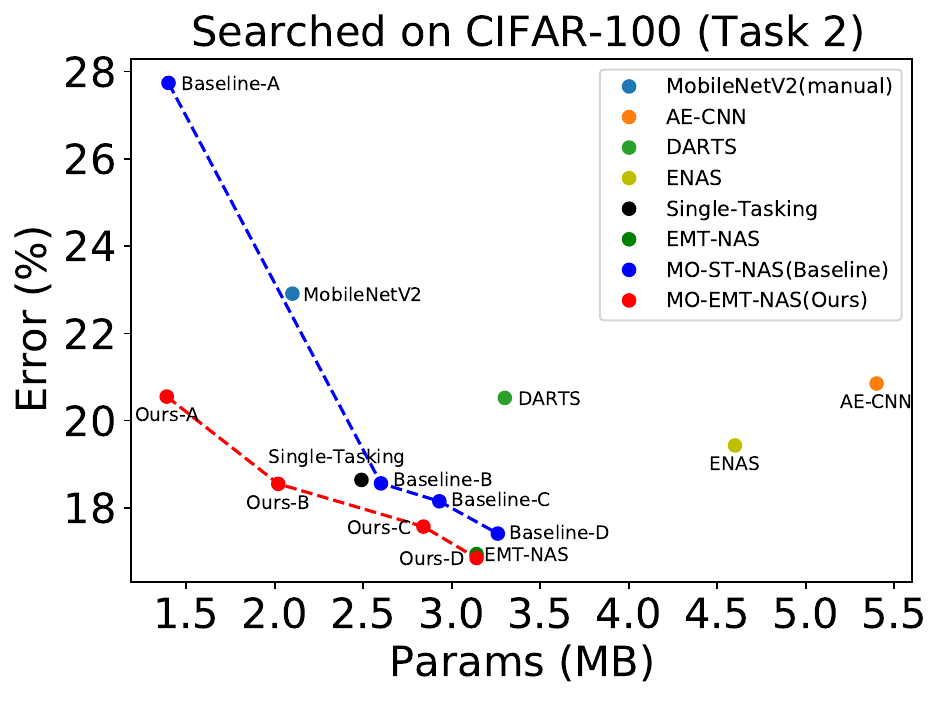}%
\label{CIFAR-100}}
\caption{Visualization of the two tasks of CIFAR-10 and CIFAR-100. The corresponding model error and size are summarized in Table A in the Supplementary Material.}
\label{fig:CIFARs}
\end{figure}

The results in Table A in the Supplementary Material show that models found by MO-EMT-NAS dominate all models found by other methods under comparison, except for Baseline-A. This indicates that MO-EMT-NAS overwhelmingly outperforms all compared approaches. As shown in Fig.~\ref{fig:CIFARs}, MO-EMT-NAS achieves a set of diverse and superior architectures (the red line is at the bottom-left). Interestingly, MO-EMT-NAS approaches are more competitive compared to single-objective MT-NAS approaches, indicating that simultaneously optimizing multiple conflicting objectives enhances the maintenance of large models without undue sacrifice of the validation error. Besides, the comparison between MO-EMT-NAS and MO-ST-NAS demonstrates that architecture knowledge transfer between tasks facilitates the search for neural architectures. The average of the HV values over five runs on CIFAR-10 and CIFAR-100 for MO-ST-NAS and MO-EMA-NAS in Table~\ref{table:CIFAR HV} further demonstrates the better convergence and diversity of our approach. It is important to highlight that the runtime of the algorithms under comparison is summarized in Table A in the Supplementary Material. Among these algorithms, MO-EMT-NAS emerges as the most efficient computationally, requiring just 0.38 GPU days for CIFAR-10 and CIFAR-100.

\begin{table}[!t]
    \centering
    \begin{minipage}[c]{0.29\linewidth}
\caption{Comparison of HV values on CIFAR-10 and CIFAR-100. A larger HV value indicates better algorithm performance in terms of convergence and diversity.}
\label{table:CIFAR HV}
\centering
\begin{tabular}{l|c}
\hline 
Approach         & CIFAR-10             \\
\hline 
Baseline  & 0.556±0.109          \\
\textbf{ours}     & \textbf{0.666±0.082}  \\
\hline 
\hline 
Approach                 & CIFAR-100       \\
\hline 
Baseline             & 0.458±0.008 \\
\textbf{ours}       & \textbf{0.579±0.225} \\
\hline 
        \end{tabular}
    \end{minipage}
    \hfill
    \begin{minipage}[c]{0.65\linewidth}
        \centering
        \includegraphics[width=0.80\linewidth]{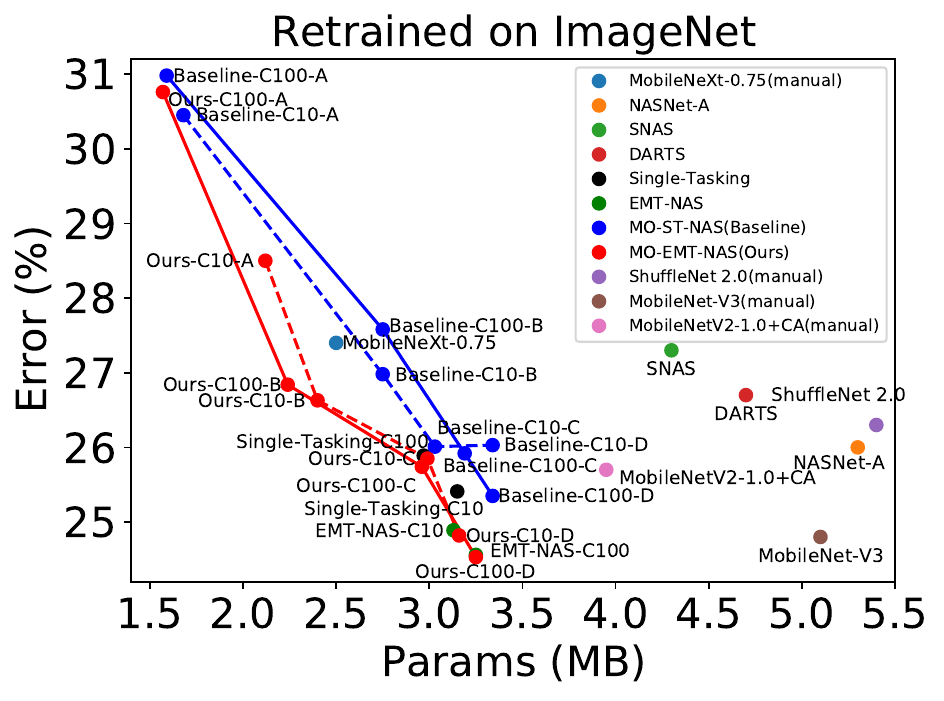}
        \captionsetup{type=figure}
        \caption{The architecture transferability tested on ImageNet by transferring  our architectures searched on CIFAR-10 and CIFAR-100 to ImageNet. }
        \label{fig:imagenet}
    \end{minipage}
\end{table}

\subsection{Transfer to ImageNet}

The 16 neural architectures obtained by MO-ST-NAS and MO-EMT-NAS on CIFAR-10 and CIFAR-100, as plotted in Fig.~\ref{fig:CIFARs}, are transferred to ImageNet for retraining. The results of the transferred architectures on ImageNet are compared with several representative algorithms and given in Table B in the Supplementary Material. From Fig.~\ref{fig:imagenet}, we observe that MO-EMT-NAS shows superior results in terms of the Top-1 accuracy than other algorithms under comparison, while providing a series of trade-off models with the number of parameters ranging from 1.57M to 3.25M. And the architectures transferred from MO-EMT-NAS always yields better performances than that from MO-ST-NAS. The model with the highest accuracy, Ours-C-100-D, has an accuracy of 75.47\% and 3.25M number of parameters. Note that the experiment on the ImageNet (a single task with a large dataset) aims to evaluate the architecture transferability of each method rather than its ability to solve multiple tasks.

\subsection{Medical Multi-Objective Multi-Tasking}

\begin{figure}[]  
\centering
\subfloat[][Two tasks (PA)]{\includegraphics[width=0.33\linewidth]{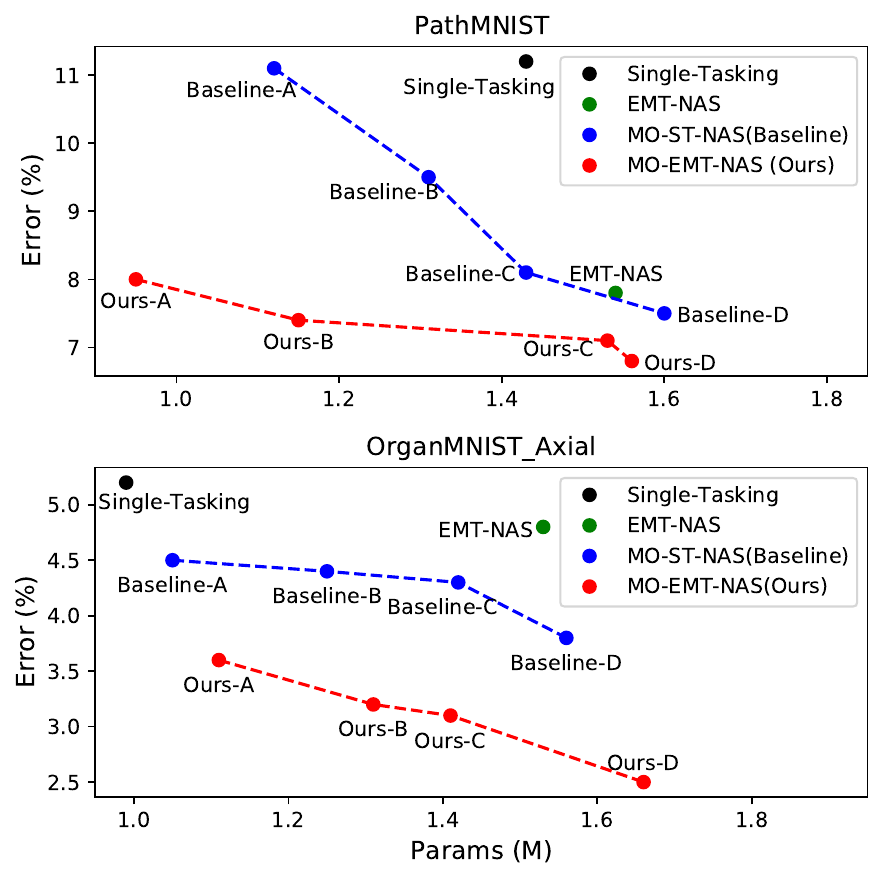}%
\label{2 tasks (PA)}}
\subfloat[][Two tasks (PC)]{\includegraphics[width=0.33\linewidth]{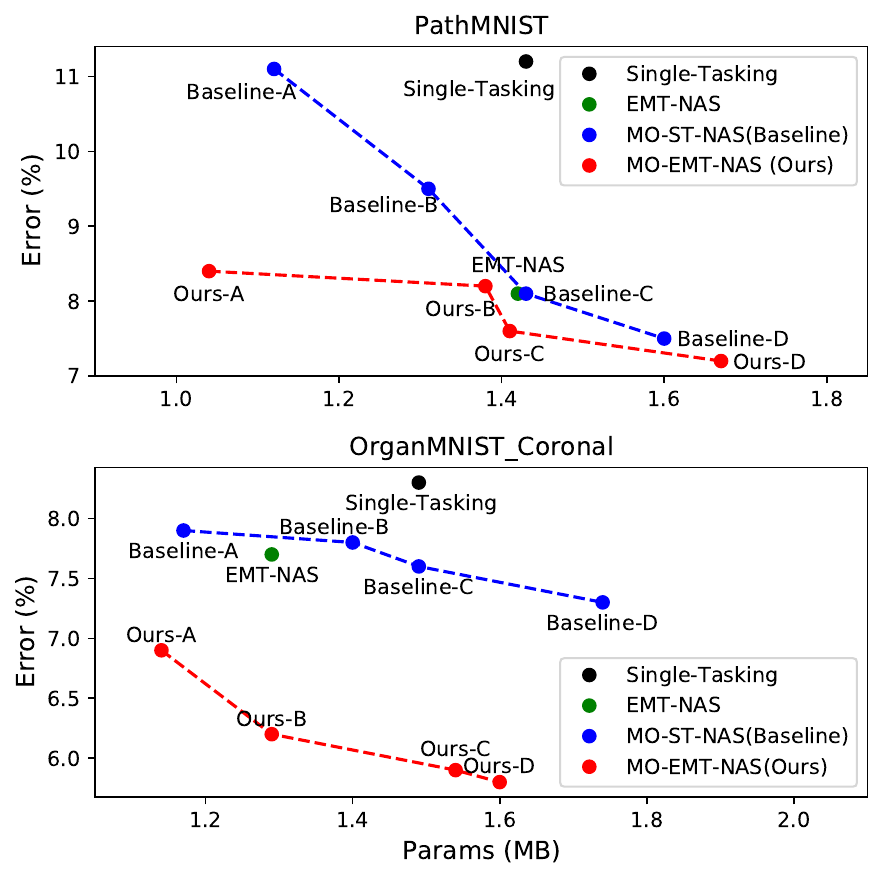}%
\label{2 tasks (PC)}}
\subfloat[][Two tasks (PS)]{\includegraphics[width=0.33\linewidth]{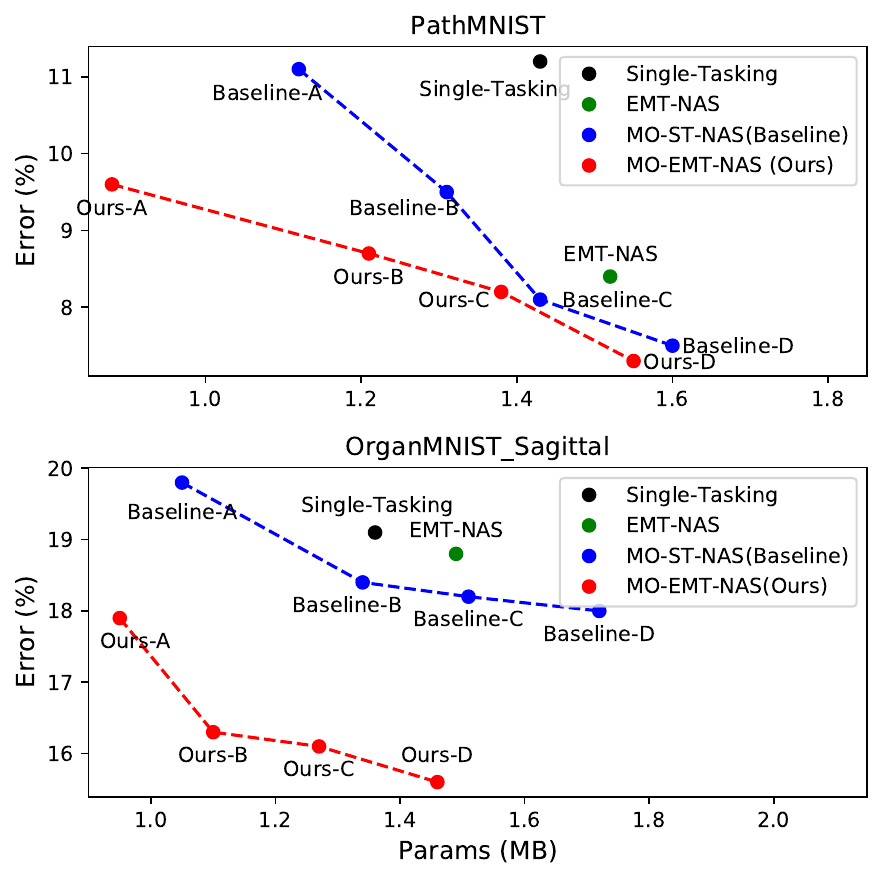}%
\label{2 tasks (PS)}}
\hfil
\subfloat[][Two tasks (AC)]{\includegraphics[width=0.33\linewidth]{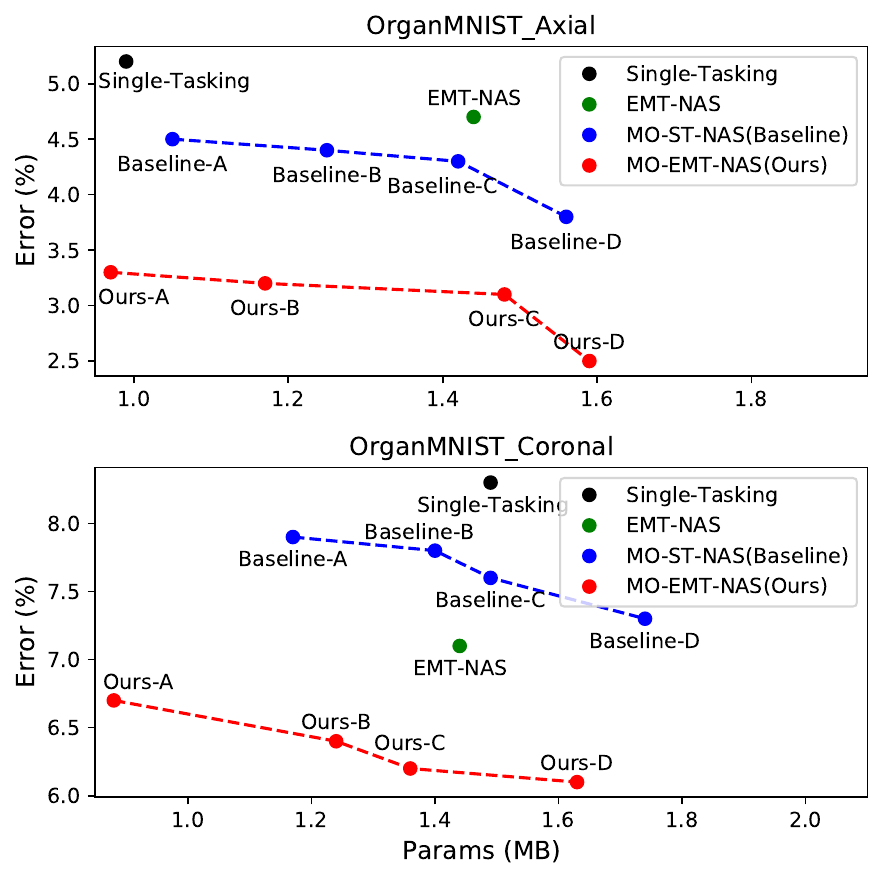}%
\label{2 tasks (AC)}}
\subfloat[][Two tasks (AS)]{\includegraphics[width=0.33\linewidth]{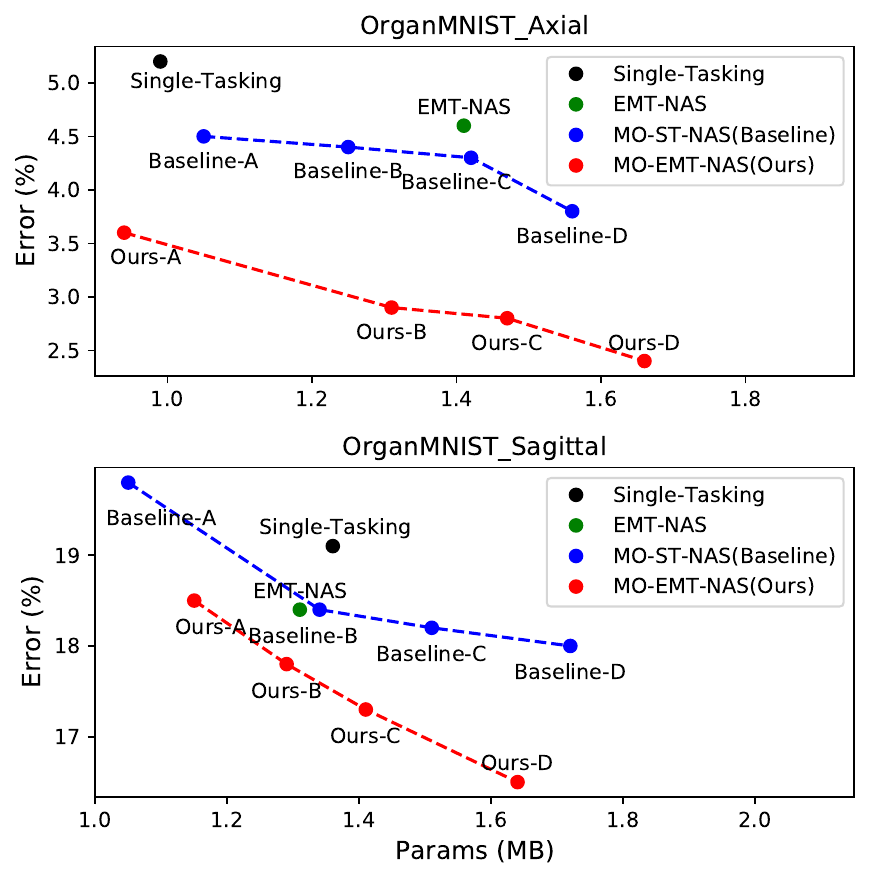}%
\label{2 tasks (AS)}}
\subfloat[][Two tasks (CS)]{\includegraphics[width=0.33\linewidth]{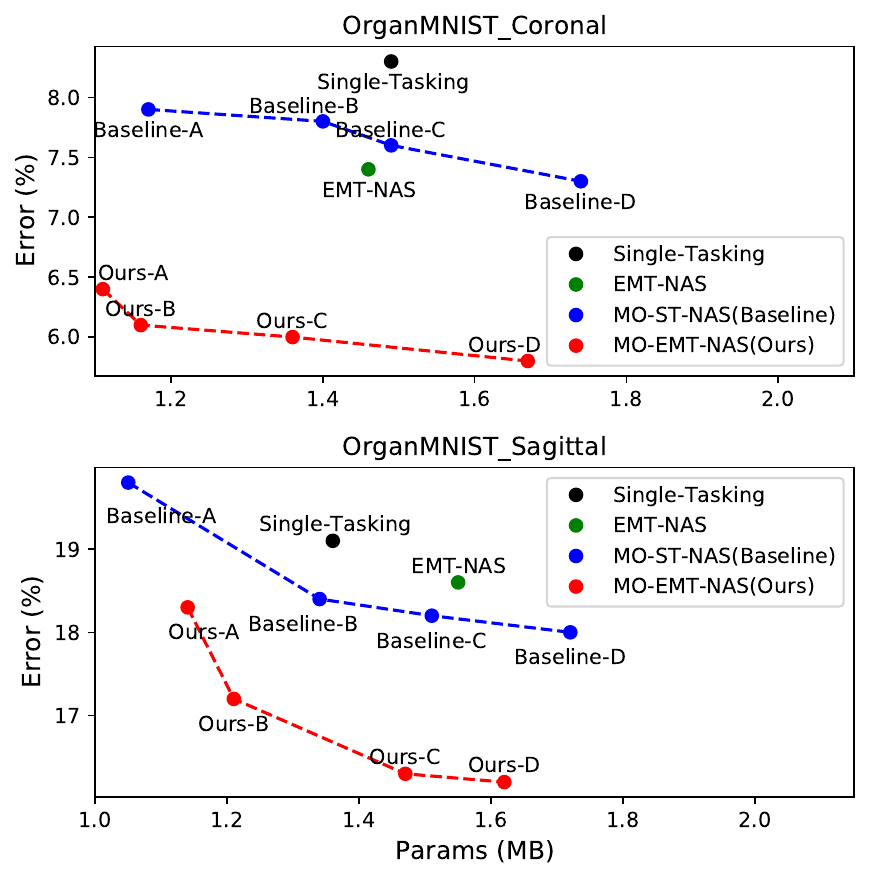}%
\label{2 tasks (CS)}}
\hfil
\subfloat[][Three  tasks (PAC)]{\includegraphics[width=0.33\linewidth]{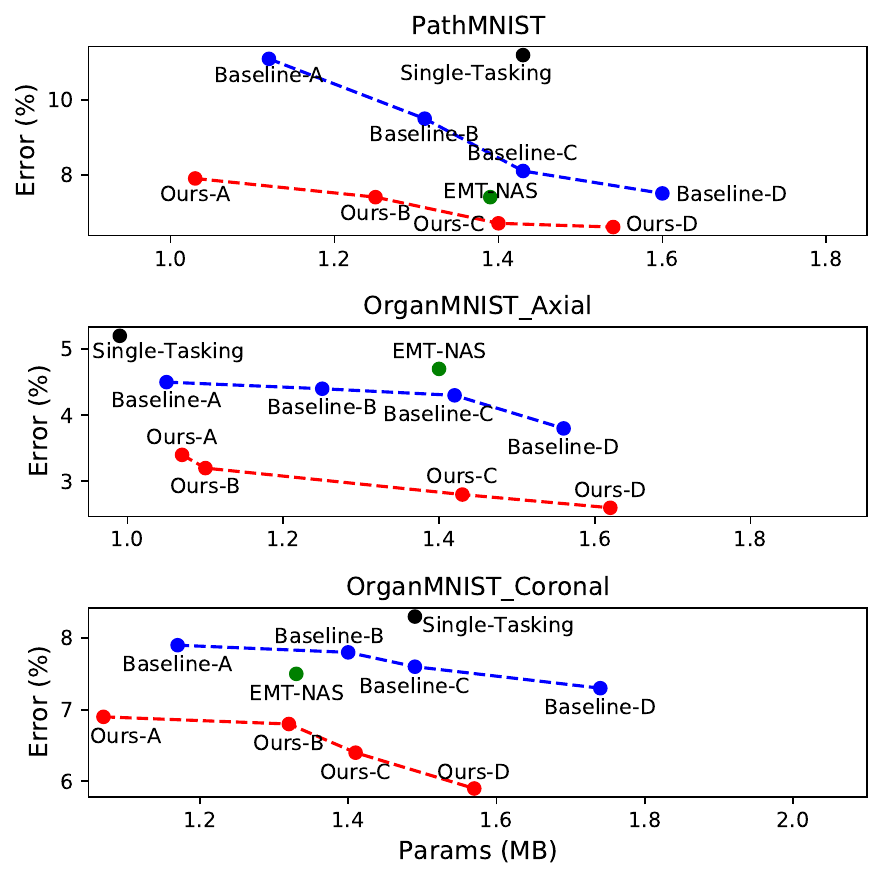}%
\label{3 tasks (PAC)}}
\subfloat[][Three  tasks (PAS)]{\includegraphics[width=0.33\linewidth]{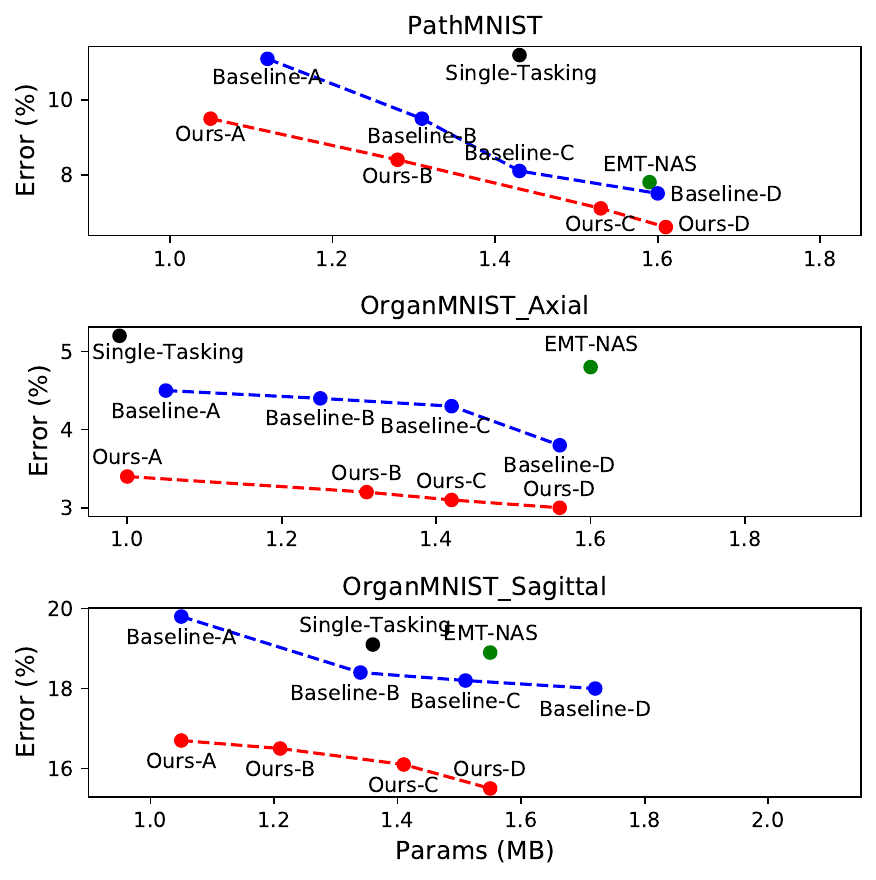}%
\label{3 tasks (PAS)}}
\subfloat[][Three  tasks (PCS)]{\includegraphics[width=0.33\linewidth]{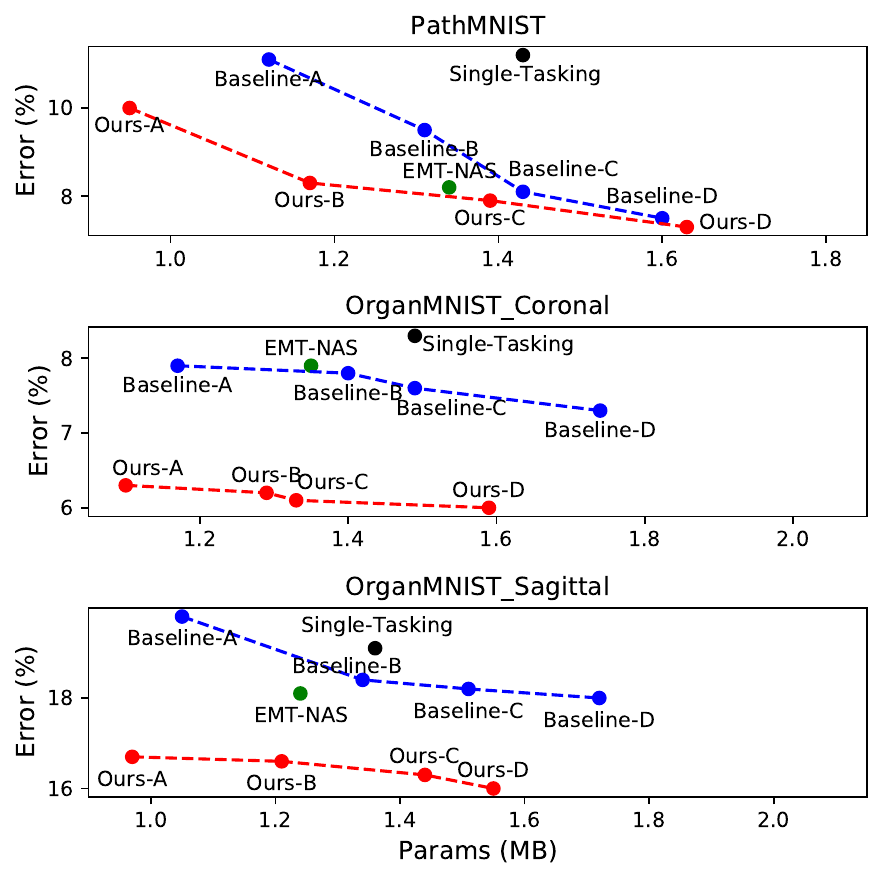}%
\label{3 tasks (PCS)}}
\hfil
\subfloat[][Three tasks (ACS)]{\includegraphics[width=0.33\linewidth]{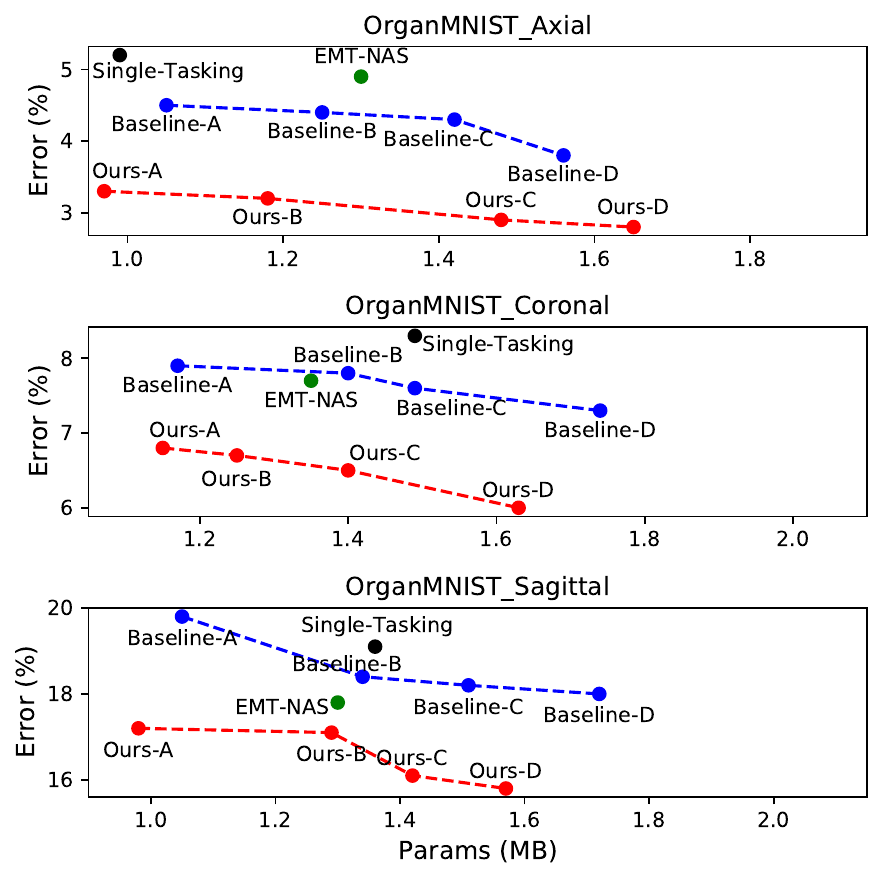}%
\label{3 tasks (ACS)}}
\subfloat[][Four tasks (PACS)]{\includegraphics[width=0.33\linewidth]{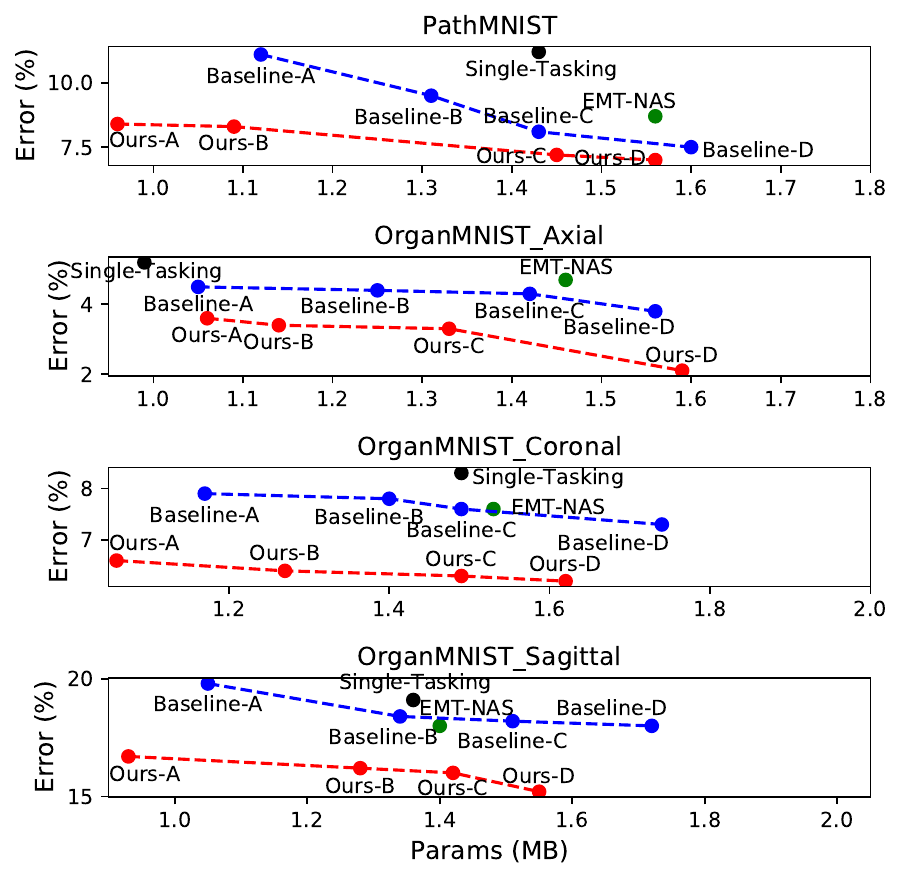}%
\label{MED 4 Tasks}}
\subfloat[c][Relatedness scores]{\includegraphics[width=0.33\linewidth]{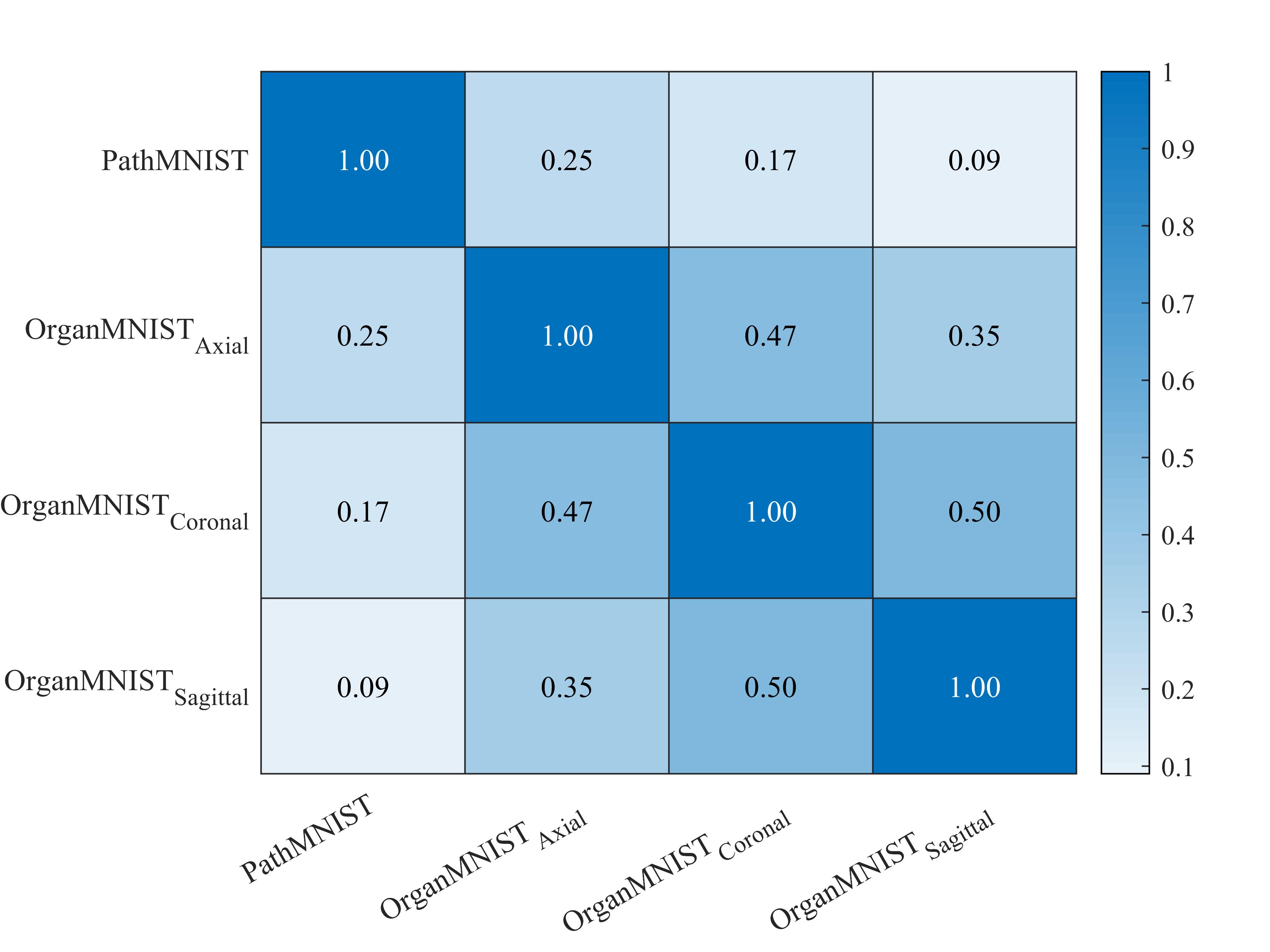}%
\label{Relatedness scores}}

\caption{Visualization on MedMNIST with two-, three- and four-tasking settings. The values of the task relatednesss score are shown in Fig. \ref{Relatedness scores}. The corresponding model error and size are summarized in Table D-N in the Supplementary Material.}
\label{MED}
\end{figure}

PathMNIST, OrganMNIST\_Axial, OrganMNIST\_Coronal, and OrganMNIST\_\\Sagittal abbreviated as P, A, C, and S.


{\noindent\bf Multi-Objective NAS:}
Figures~\ref{2 tasks (PA)}-\ref{MED 4 Tasks} show that MO enables MO-EMT-NAS to yield a set of promising models with respect to the accuracy, model size or both. This further confirms the advantage of using the MO methods with the auxiliary objective. Importantly, MO-EMT-NAS finds a set of neural architectures with a low error that dominate the single models found by both single-objective NAS architectures, the Single-Tasking and EMT-NAS.

{\noindent\bf Evolutionary Multi-Tasking NAS:}
In Table~\ref{table:MED 2}, the obtained Pareto optimal architecture set for each task is evaluated by the HV metric. Compared with MO-ST-NAS, MO-EMT-NAS achieves higher HV values, i.e., better performance in terms of convergence and diversity, on various task combinations. This is accomplished by using the knowledge transfer across tasks to promote the multi-tasking optimization. Across all settings, MO-EMT-NAS consistently achieves better accuracy while being significantly faster than MO-ST-NAS.

{\noindent\bf Scalability of MO-EMT-NAS:}
The scalability of MO-EMT-NAS is tested by setting the number of tasks to two, three, and four, respectively. As illustrated in Figs.~\ref{2 tasks (PA)}-\ref{MED 4 Tasks}, MO-EMT-NAS consistently exhibits superior performance compared to single-objective NAS approaches, confirming the promising scalability of MO-EMT-NAS. Specifically, architectures discovered by MO-EMT-NAS consistently dominate (with better performance) or are non-dominated (with similar performance) compared to those found by EMT-NAS and Single-Tasking NAS.

{\noindent\bf Multitasking with Different Similarities:}
Using ResNet-50 as a feature extractor, we conduct the representation similarity analysis \cite{dwivedi2019representation} to obtain task relatedness scores (RS) \cite{ben2008notion} between the four medical datasets. The RS results presented in Fig.~\ref{Relatedness scores} vary from 0.09 to 0.50. Notably, one can observe lower scores between P and A, C, S and higher scores between A, C, S. According to Fig.~\ref{2 tasks (PA)}-\ref{2 tasks (PS)}, a performance drop in terms of the error can be observed with the decrease of the RS. For example, MO-EMT-NAS finds a set of Pareto optimal models with errors ranging from 6.8\% to 8.0\% on the dataset P for the two tasks with $RS=0.25$ in Fig. \ref{2 tasks (PA)} while obtaining models with errors ranging from 7.3\% to 9.6\% on the dataset P in the two tasks PS with $RS=0.09$ in Fig. \ref{2 tasks (PS)}. A possible reason is that the lack of similarity between tasks poses challenges in architectural knowledge transfer, since less transferable information can be obtained.


\begin{table}[!t]
\scriptsize
\caption{Comparison of HV values and runtime on MedMNIST.}
\label{table:MED 2}
\centering
\begin{tabular}{c|c|cccc|cccc}
\hline 
\multirow{3}{*}{\# Tasks}
&\multirow{3}{*}{Model}   
& \multirow{3}{*}{\begin{tabular}[c]{@{}c@{}}{Path} \\HV\\ $\uparrow$ \end{tabular}}
& \multirow{3}{*}{\begin{tabular}[c]{@{}c@{}}{Organ\_A} \\HV\\ $\uparrow$ \end{tabular}}
& \multirow{3}{*}{\begin{tabular}[c]{@{}c@{}}{Organ\_C} \\HV\\ $\uparrow$ \end{tabular}}
& \multirow{3}{*}{\begin{tabular}[c|]{@{}c@{}}{Organ\_S} \\HV\\ $\uparrow$ \end{tabular}}
& \multirow{3}{*}{\begin{tabular}[|c]{@{}c@{}}{GPU} \\ Days \Rmnum{1}\\{(\%) $\downarrow$} \end{tabular}} 
& \multirow{3}{*}{\begin{tabular}[c]{@{}c@{}}{GPU} \\ Days \Rmnum{2}\\{(\%) $\downarrow$} \end{tabular}} 
& \multirow{3}{*}{\begin{tabular}[c]{@{}c@{}}{Time} \\  \Rmnum{1}\\{(\%) $\downarrow$} \end{tabular}}
& \multirow{3}{*}{\begin{tabular}[c]{@{}c@{}}{Time} \\  \Rmnum{2}\\{(\%) $\downarrow$} \end{tabular}}  \\ &&&&&&\\&&&&& \\

\hline 
1&Baseline 
& 0.47±0.04 & 0.45±0.08 & 0.62±0.13 & 0.49±0.06 & +00.0  & +00.0 &  +00.0  &  +00.0 \\
\hline 
2&Ours
& 0.72±0.07& 0.66±0.05 & & &-59.7      & -32.2 & -26.5 & -50.8 \\
2&Ours  
& 0.69±0.08& & 0.69±0.05 & & -62.8     & -44.3 & -40.1 & -50.1   \\
2&Ours   
& 0.75±0.09 & & & 0.64±0.08 & -62.9     & -43.7 & -39.7 & -49.7  \\
2&Ours    
&& 0.66±0.07& 0.67±0.12&  & -68.4     & -42.1 & -35.9 & -50.3  \\
2&Ours       
&& 0.69±0.05&& 0.63±0.06  & -68.6     & -42.0 & -36.0 & -49.5  \\
2&Ours     
&&& 0.66±0.07& 0.56±0.03  & -76.5     & \textbf{-53.5} & -49.1 & -49.8  \\
\hline 
3&Ours     
& 0.76±0.05& \textbf{0.75±0.09}& 0.72±0.09    &   & -70.8     & -41.4 & -50.8 & -66.3  \\
3&Ours       
& 0.71±0.06& 0.67±0.05&& 0.64±0.07& -70.6     & -40.5 & -50.2 & -65.9  \\
3&Ours     
& 0.76±0.11&& \textbf{0.74±0.11}& \textbf{0.72±0.11} & -73.6     & -50.4 & -55.8  & -66.1 \\
3&Ours    
&& 0.69±0.06& 0.67±0.07& 0.70±0.10  &\textbf{-77.7}     & -48.6 & \textbf{-60.9} & -66.9  \\
\hline 
4&Ours     
& \textbf{0.77±0.13}& 0.68±0.09& 0.72±0.11& 0.72±0.13    & -76.4    & -44.3& -52.1 & \textbf{-74.7}  \\

\hline 
\end{tabular}
\end{table}

{\noindent\bf Search Efficiency:}
The runtime during the experiments is recorded and the percentage of the saved time by each algorithm compared with MO-ST-NAS is measured and presented in Table~\ref{table:MED 2}. 

The time saved by MO-EMT-NAS with addressing tasks in parallel is denoted as "GPU Days \Rmnum{1} (\%)". One can observe that compared with the multi-objective single-tasking baseline, the proposed MO-EMT-NAS reduces the runtime from 59.7\% to 77.7\% for jointly addressing two, three and four tasks, while reaching a better balance between the error and model size. Besides, the time saved by MO-EMT-NAS unsurprisingly increases with the increase of the number of tasks solved jointly. The main reason is that the parallel training and evaluation of multiple tasks in MO-EMT-NAS significantly improves the computational efficiency and caps the overall runtime to that of the slowest task. Hence, for two-task settings, the time saving does not exceed 50\%. Besides, while MO-EMT-NAS handles multiple tasks simultaneously, MO-ST-NAS solves tasks one by one, resulting in much more computational cost. 

To further investigate the efficiency of MO-EMT-NAS, the time reduced by MO-EMT-NAS without the parallelization of the training and evaluation on multiple tasks is denoted as "GPU Days \Rmnum{2} (\%)". More specifically, the time saved for training and validation and that for searching are measured and denoted as "Time \Rmnum{1} (\%)" and "Time \Rmnum{2} (\%)", respectively. The results of "GPU Days \Rmnum{2} (\%)" show that MO-EMT-NAS obtains up to 53.5\% time savings compared with MO-ST-NAS, indicating the efficiency gained from the multi-tasking framework. Interestingly, by comparing "GPU Days \Rmnum{1} (\%)" and "GPU Days \Rmnum{2} (\%)", we can confirm the existence of heterogeneous time costs of different tasks. According to "Time \Rmnum{1} (\%)", MO-EMT-NAS reduces up to 60.9\% the time cost for training and validation. Similarly, the time spent on searching using an evolutionary algorithm is significantly reduced with the increase of the number of tasks. The reason is that the EA requires almost the same time for each task, accordingly the time for searching is doubled if addressing tasks one by one.

\subsection{Ablation studies}

\begin{table}[!t]
\scriptsize
    \centering
    \begin{minipage}[c]{0.40\linewidth}
\caption{Comparison of CIFAR-10 and CIFAR-100 with/without the auxiliary objective. A larger HV value indicates better algorithm performance in terms of convergence and diversity.}
\label{table:HV}
\centering
\begin{tabular}{ccc}
\hline 
Approach         & CIFAR-10        & CIFAR-100    \\
\hline 
without  & 0.443±0.058 & 0.356±0.004 \\
\textbf{with}  & \textbf{0.571±0.068} & \textbf{0.532±0.138} \\
\hline 
\end{tabular}
    \end{minipage}
    \hfill
    \begin{minipage}[c]{0.55\linewidth}
\caption{Effect of the random mating probability on performance on PathMNIST (Path) and OrganMNIST\_Axial (Organ\_A).}
\label{table:RMP}
\centering
\begin{tabular}{ccc}
\hline 
RMP         & Path        & Organ\_A      \\
\hline 
0.0  & 0.731±0.114 & 0.641±0.054 \\
0.2  & 0.733±0.144 & 0.692±0.105 \\
0.4  & 0.784±0.049 & 0.649±0.050 \\
0.6  & 0.788±0.068 & 0.590±0.122 \\
0.8  & 0.821±0.092 & 0.710±0.144 \\
\textbf{1.0}  & \textbf{0.823±0.077} & \textbf{0.718±0.079} \\

\hline 
\end{tabular}
    \end{minipage}
    
\end{table}

To validate the efficiency of the auxiliary objective, MO-EMT-NAS with and without the auxiliary objective $f_a$ are performed on CIFAR-10 and CIFAR-100. The fact that the HV values achieved by MO-EMT-NAS, i.e., 0.571 for CIFAR-10 and 0.532 for CIFAR-100, are better than that achieved by MO-EMT-NAS without $f_a$, i.e., 0.443 for CIFAR-100 and 0.356 for CIFAR-100, convincingly showcasing the advantage of using $f_a$. The HV values in Table~\ref{table:HV} indicate that MO-EMT-NAS with the help of the proposed auxiliary objective yields a set of well-converged and diverse non-dominated architectures.

\subsection{Sensitivity Analysis}

The random mating probability (RMP) is an important parameter that controls the degree of knowledge transfer between tasks. Hence, $RMP$ is set to 0, 0.2, 0.4, 0.6, 0.8 and 1 to test its impact on the performance, and the HV results on two datasets, Path and Organ\_A, are summarized in Table~\ref{table:RMP}. Accordingly, we find that MO-EMT-NAS with a higher $RMP$ value tends to achieve better performance on Path, shedding lights on the potential advantage of encouraging the architectural knowledge transfer. Indeed, a higher degree of knowledge transfer indicated by a larger $RMP$ does not always improve the performance on Organ\_A, but the best performance is achieved when $RMP=1$. 


We have also extensively tuned the crossover and mutation probabilities, population size and the number of generations. The results in terms of HV values are presented in Supplementary Material D.

\section{Conclusion}
\label{sec:conclusion}

In this work, we propose a multi-objective multi-tasking NAS framework with the help of weight-sharing-based supernets to efficiently achieve a set of promising architectures with diverse model sizes. The multi-tasking framework enables architecture knowledge acquired from different tasks to be implicitly transferred, thereby effectively solving multiple tasks from different datasets. To mitigate the small model trap problem, we introduce an auxiliary objective that prefers larger models over smaller ones when they achieve similar accuracy, thereby achieving a set of promising architectures with various model sizes. Extensive experiments demonstrate that the architectures obtained by MO-EMT-NAS exhibit superior performance at a lower computational cost than the state of the art while being able to maintain a high degree of diversity in model sizes. 

\section*{Acknowledgements}
This work was supported by National Natural Science Foundation of China (Key Program: 62136003), the Shanghai Committee of Science and Technology, China (Grant No.22DZ1101500), Fundamental Research Funds for the Central Universities (222202417006) and the Programme of Introducing Talents of Discipline to Universities (the 111 Project) under Grant B17017 and Shanghai AI Lab.

%
%
\bibliographystyle{splncs04}
\bibliography{main}
\end{document}